\documentclass[sigconf, screen]{acmart}

\usepackage{multirow}
\usepackage{float}
\usepackage{stfloats}
\usepackage{balance}

\copyrightyear{2023}
\acmYear{2023}
\setcopyright{acmlicensed}\acmConference[MMSports '23]{Proceedings of the 6th International Workshop on Multimedia Content Analysis in Sports}{October 29, 2023}{Ottawa, ON, Canada}
\acmBooktitle{Proceedings of the 6th International Workshop on Multimedia Content Analysis in Sports (MMSports '23), October 29, 2023, Ottawa, ON, Canada}
\acmPrice{15.00}
\acmDOI{10.1145/3606038.3616158}
\acmISBN{979-8-4007-0269-3/23/10}

\acmSubmissionID{mmsp015}

\begin{document}

\title{Dynamic NeRFs for Soccer Scenes}

\author{Sacha Lewin}
\affiliation{
  \institution{University of Li\`ege}
  \city{Li\`ege}
  \country{Belgium}
}
\email{sacha.lewin@uliege.be}

\author{Maxime Vandegar}
\affiliation{
  \institution{EVS Broadcast Equipment}
  \city{Li\`ege}
  \country{Belgium}
}
\email{m.vandegar@evs.com}

\author{Thomas Hoyoux}
\affiliation{
  \institution{EVS Broadcast Equipment}
  \city{Li\`ege}
  \country{Belgium}
}
\email{t.hoyoux@evs.com}

\author{Olivier Barnich}
\affiliation{
  \institution{EVS Broadcast Equipment}
  \city{Li\`ege}
  \country{Belgium}
}
\email{o.barnich@evs.com}

\author{Gilles Louppe}
\affiliation{
  \institution{University of Li\`ege}
  \city{Li\`ege}
  \country{Belgium}
}
\email{g.louppe@uliege.be}

\begin{abstract}

The long-standing problem of novel view synthesis has many applications, notably in sports broadcasting. Photorealistic novel view synthesis of soccer actions, in particular, is of enormous interest to the broadcast industry. Yet only a few industrial solutions have been proposed, and even fewer that achieve near-broadcast quality of the synthetic replays. Except for their setup of multiple static cameras around the playfield, the best proprietary systems disclose close to no information about their inner workings. Leveraging multiple static cameras for such a task indeed presents a challenge rarely tackled in the literature, for a lack of public datasets: the reconstruction of a large-scale, mostly static environment, with small, fast-moving elements. Recently, the emergence of neural radiance fields has induced stunning progress in many novel view synthesis applications, leveraging deep learning principles to produce photorealistic results in the most challenging settings. In this work, we investigate the feasibility of basing a solution to the task on \textit{dynamic NeRFs}, i.e., neural models purposed to reconstruct general dynamic content. We compose synthetic soccer environments and conduct multiple experiments using them, identifying key components that help reconstruct soccer scenes with dynamic NeRFs. We show that, although this approach cannot fully meet the quality requirements for the target application, it suggests promising avenues toward a cost-efficient, automatic solution. We also make our work dataset and code publicly available, with the goal to encourage further efforts from the research community on the task of novel view synthesis for dynamic soccer scenes. For code, data, and video results, please see \url{https://soccernerfs.isach.be}.

\end{abstract}



\begin{teaserfigure}
  \includegraphics[width=\textwidth]{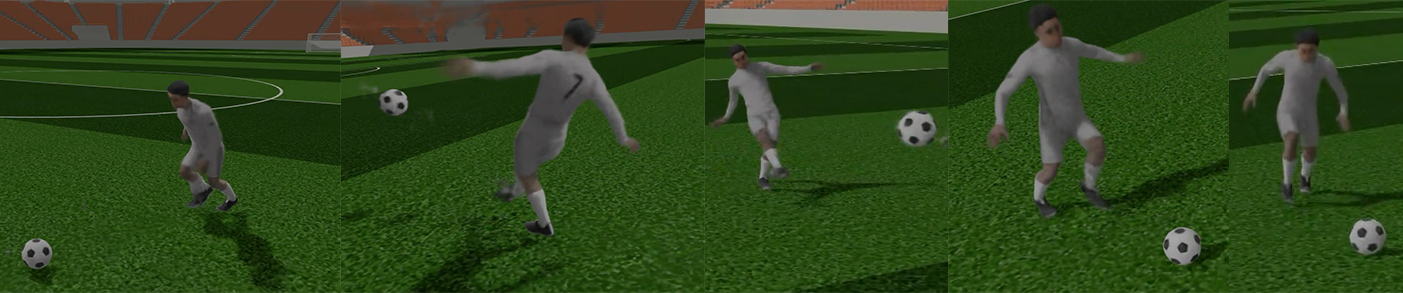}
  \caption{Novel view synthesis in a synthetic dynamic environment, given 30 known views and camera poses.}
  \Description{Five images of a well-modeled synthetic soccer player in white interacting with a ball at the center of a field.}
  \label{fig:teaser}
\end{teaserfigure}

\maketitle

\section{Introduction}

Synthesizing novel views of a scene from a sparse sample of images is a long-standing problem in computer vision \cite{ViewInterpolationImage_chen1993, LightFieldRendering_levoy1996, PhotorealisticSceneReconstruction_seitz1997}. A notable field of application is sports broadcasting, in which action replays have a major role in story-telling and performance analysis. As one of the most popular sports, soccer receives a lot of broadcast coverage from top to low-tier competitions all over the world, with much care given to making the viewer experience ever more pleasant and engaging. Augmenting the broadcast production of soccer events with novel-view video synthesis of action replays is therefore very attractive to industrial actors, and a real opportunity for the computer vision research community.

Despite the industry interest in novel view synthesis of soccer replays, only a few proprietary systems exist on the market. Indeed, such interest cannot outweigh the need for the highest image quality in broadcast productions; the industry, therefore, imposes very high standards in terms of the photorealism of the synthesized views. One noteworthy system \cite{IntelTrueView} is able to deliver synthetic replays that are stunningly photorealistic, but for a few visual artifacts. Their setup is composed of dozens of very high-resolution static cameras, installed all around the soccer field high up above the bleachers. Their image data are processed by private, proprietary software running on very powerful hardware. These image data remain private as well, and equivalent public datasets are simply nonexistent. The only insight offered by this system to the research community is the validity of using a static multi-camera setup for the task.

Even with no image data available, one can reason about the challenges that arise from using an array of distant static cameras as a basis for the reconstruction of a soccer environment. Outdoor sports like soccer are composed of a large static environment, the stadium, and small dynamic elements, the players and the ball. Traditional computer vision methods would most likely have to rely on very high-resolution images, as in \cite{IntelTrueView}, to reconstruct an underlying 3D model of the scene able to faithfully render the movements of the small dynamic elements. Having to deal with massive amounts of image data for reconstructing a single, short soccer action is however not a desirable property for a solution.

Building on the modern deep learning-based paradigm to computer vision problems, neural radiance fields \cite{NerfRepresentingScenes_mildenhall2021} (NeRFs) have recently become the state of the art for high-quality novel view synthesis, and have been widely improved and extended to produce excellent results in very challenging settings. A notable line of work is \textit{dynamic} NeRFs, i.e. neural models purposed to reconstruct spatiotemporal content, as opposed to only spatial, static content. This, therefore, begs the question: \textit{Are dynamic NeRFs suitable for reconstructing soccer scenes?} To find potential answers to this question, we propose this exploratory work, in which we make three important assumptions.

First, we only consider camera setups similar to the one used by the aforementioned proprietary system \cite{IntelTrueView}, deeming it optimal for the task at hand. Specifically, we use an array of 20 to 30 static cameras, positioned all around the soccer stadium and pointing toward the soccer field. This assumption goes well with the working conditions usually recommended to achieve good performance with NeRFs. Moreover, most NeRFs assume input views to be calibrated by third-party Structure from Motion (SfM) tools, which are known to bring robust results with such camera setups in mostly static environments, such as a soccer stadium.

Second, we limit our study to synthetic soccer datasets, yet we believe its results also apply to real data. As already mentioned, soccer image datasets with the considered camera setups are virtually nonexistent for the public, to the best of our knowledge. We therefore composed synthetic datasets, using public computer graphics engines and models. Because we control the cameras in our 3D virtual environments, this assumption also allows us to leave camera calibration aspects out of the scope of our work. We are confident that our findings remain valid when working on real use cases, given the availability of robust SfM tools, and the reputation of very good photorealism of NeRFs with real image data.

Third, we only consider \textit{general} dynamic NeRFs, i.e., dynamic NeRFs with \textit{no domain knowledge}, to identify early limitations of the neural-based reconstruction paradigm in the context of our task. Another important reason is that domain-specific priors are often difficult and expensive to produce. For instance, an accurate skeletal reconstruction of the players would be predictably very useful for soccer replay synthesis, but is a hard task in itself, especially with the considered camera setups. Our goal is to avoid resorting to such priors, which are likely to be complex and costly. This assumption also has the advantage to make our study potentially insightful for the use of dynamic NeRFs for other sports than soccer, given similar camera setups.

We select recent state-of-the-art general dynamic NeRF models and compare them in three synthetic soccer environments of increasing complexity. Our aim is to progressively transition from ideal conditions for the considered models, to conditions that are similar to the optimal camera setup used in \cite{IntelTrueView}.

Our contributions could be summarized as follows:
\begin{enumerate}
    \item We provide a study of the performance of general dynamic NeRFs on the task of soccer replay synthesis in increasingly complex environments. Models are studied as they were introduced in the literature, then augmented with general, non-domain specific components that we identify. We close the study with a higher-level discussion about limitations and future work.
    \item As we wish to foster research efforts toward solving this challenging task, we publicly release our code, including the improving components and experimental settings, and our complete work dataset, including images, depth maps, Blender \cite{Blender3DModelling_community2018} scripts, and camera calibrations for all synthetic environments. These are all ready-to-use in Nerfstudio \cite{NerfstudioModularFramework_tancik2023}, a rich and popular open-source framework for using and developing NeRF models.
\end{enumerate}

The remainder of this paper is organized as follows. Section~\ref{background} provides preliminaries about NeRFs and introduces their extension to dynamic environments. Section~\ref{implem} details our experimental setup: methods, evaluation, and environments. Section~\ref{experiments} showcases and discusses results. Finally, Section~\ref{discussion} provides a higher-level discussion about the feasibility of using these methods, along with some paths for improvement.

\section{Neural Scene Representation}\label{background}

\paragraph{Neural Radiance Fields \cite{NerfRepresentingScenes_mildenhall2021}}

The original neural radiance field (NeRF) model implicitly encodes a scene in the weights of a multi-layer perceptron (MLP). The model learns to associate density and color information to any point in space, which allows for rendering images using classical volume rendering \cite{RayTracingVolume_kajiya1984, LightDiffusionClouds_max1986}. This process is end-to-end differentiable, which allows for training using only captured views and their associated camera poses.
For improving training and rendering time, more recent methods \cite{InstantNeuralGraphics_muller2022, TensoRFTensorialRadiance_chen2022, Sun2021DirectVG} use a hybrid approach, leveraging both implicit and explicit representations, such as voxel grids. Those methods store learnable features, which are then decoded with an MLP. 

\paragraph{Dynamic NeRFs}\label{dyn_nerfs}
Various techniques have been proposed to extend NeRFs to dynamic reconstruction. Methods such as \cite{DNeRFNeuralRadiance_pumarola2021, NeuralRadianceFlow_du2021, NerfiesDeformableNeural_park2021} learn a separate field, known as a deformation network, that maps each point to its corresponding position in a canonical scene. Other methods input time to the radiance field. While direct conditioning on time provides poor results \cite{DNeRFNeuralRadiance_pumarola2021}, indirect conditioning \cite{Neural3DVideo_li2022, NeRFPlayerStreamableDynamic_song2023, KPlanesExplicitRadiance_fridovich-keil2023, MixedNeuralVoxels_wang2022, FastDynamicRadiance_fang2022, HyperReelHighFidelity6DoF_attal2023} obtains state-of-the-art results on various popular benchmarks. Some models leverage domain knowledge, such as Human NeRFs. They often work by learning the motion of a skinned multi-person linear model (SMPL \cite{SMPLSkinnedMultiperson_loper2015}) along with its appearance \cite{NeuralBodyImplicit_peng2020, AnimatableNeuralRadiance_peng2021, HumanNeRFEfficientlyGenerated_zhao2022}. A more recent method supports human-object interactions \cite{HOSNeRFDynamicHumanobjectscene_liu2023}. These specific models still require complex and controlled setups. Two non domain-specific models, K-planes and NeRFPlayer, are of particular interest to us, based on their state-of-the-art performance on diverse benchmarks, and the approach they take to the reconstruction of dynamic content.

\paragraph{K-Planes \cite{KPlanesExplicitRadiance_fridovich-keil2023}}
This model builds upon methods \cite{TensoRFTensorialRadiance_chen2022, Tensor4DEfficientNeural_shao2023} that factorize the 4D space into 6 planes, corresponding to each pair of coordinates. This approach, and concurrent work \cite{EfficientGeometryAware3D_chan2022, Wadhwani2022SqueezeNeRFFF}, offer greatly-improved efficiency with high-quality results. The planes store feature vectors uniformly in space and time, at increasing scales, similar to the multiresolution hash encoding used in \cite{InstantNeuralGraphics_muller2022}. The feature vectors associated with a given point in space and time are then decoded by a shallow MLP into a density and an RGB color. K-Planes reaches state-of-the-art performance on various datasets.

\paragraph{NeRFPlayer \cite{NeRFPlayerStreamableDynamic_song2023}} This method introduces two main contributions: (i) a dynamic version of traditional explicit feature storage, such as the hash encoding from \cite{InstantNeuralGraphics_muller2022}, by using a sliding window over a larger fixed-size feature vector, and (ii) a scene decomposition into different areas depending on their nature: static, deformed, or new. Each area is modeled with a different approach, which is mostly beneficial to monocular setups. On common dynamic multi-view datasets \cite{Neural3DVideo_li2022}, NeRFPlayer reaches high-quality results, similar to K-Planes.

\section{Implementation}\label{implem}

The selected methods are K-Planes \cite{KPlanesExplicitRadiance_fridovich-keil2023} and NeRFPlayer \cite{NeRFPlayerStreamableDynamic_song2023}, outlined in Section~\ref{dyn_nerfs}. These versions are implemented in Nerfstudio \cite{NerfstudioModularFramework_tancik2023}, an open-source framework that we use for all our experiments. For fair comparisons, shared settings between the models are identical, such as proposal sampling and scene contraction \cite{MipNeRF360Unbounded_barron2022}. Model-specific hyperparameters follow the original implementations, except for the model size. We increase the hash map size of NeRFPlayer to $2^{20}$ with a temporal dimension of 64 and use Nerfacto \cite{NerfstudioModularFramework_tancik2023} as the backbone. We also drop the decomposition from NeRFPlayer, which mainly benefits monocular setups and results in unnecessarily large models. We add two additional scales to K-Planes, resulting in multiscale resolutions from $2^6$ to $2^{11}$. When enabled, ray importance sampling based on global medians (ISG) is employed \cite{Neural3DVideo_li2022}. Training follows typical Nerfstudio settings: models are trained for 30,000 iterations using Adam \cite{AdamMethodStochastic_kingma2014} with a learning rate of $10^{-2}$, which takes about 1 to 2 hours on an NVIDIA RTX 3090 GPU for each scene. Unlike typical methods which train using downsampled images for faster training, we observe improvements when using full-resolution 1080p images in our environments, without large increases in training time.

We make both our code and datasets publicly available. The former includes slightly modified versions of K-Planes and NeRFPlayer, more convenient data management for dynamic environments, training settings, and other components mentioned in Section~\ref{experiments}, such as ray importance sampling and dedicated metrics. The latter include training images, calibrated poses, depth maps, Blender files, and data parsers to readily conduct experiments within Nerfstudio.

\subsection{Evaluation}

Three metrics are typically used for assessing the visual quality of novel view synthesis: (i) PSNR, which computes differences at the pixel level, (ii) SSIM \cite{ImageQualityAssessment_wang2004}, which takes structural changes into account, and (iii) LPIPS \cite{UnreasonableEffectivenessDeep_zhang2018}, based on features in deep convolutional networks which better correlate with human judgment. Quantitative evaluation is known to be a difficult task in novel view synthesis applications and, sometimes, to hardly reflect visual quality accurately. Environments like ours make it even more challenging. Indeed, the dynamic content of interest is the players and the ball, which occupy a small region of the images. As the metrics are computed over the whole image, they are barely affected by the reconstruction quality of small elements of interest. Furthermore, we consider dynamic scenes, and computing per-frame metrics conveys no information about the temporal consistency of the results.

While the first issue can be tackled in synthetic environments by including additional views close to the content of interest, it is often not possible in real conditions. To address this, we propose alternative versions of these three metrics which are computed in restricted bounding boxes around the dynamic content. The boxes can be automatically generated by simply using an object detection model such as RetinaNet \cite{FocalLossDense_lin2017}. We refer to them as \textit{focused metrics}.

\begin{figure*}[!ht]
	\centering
	\includegraphics[width=\textwidth]{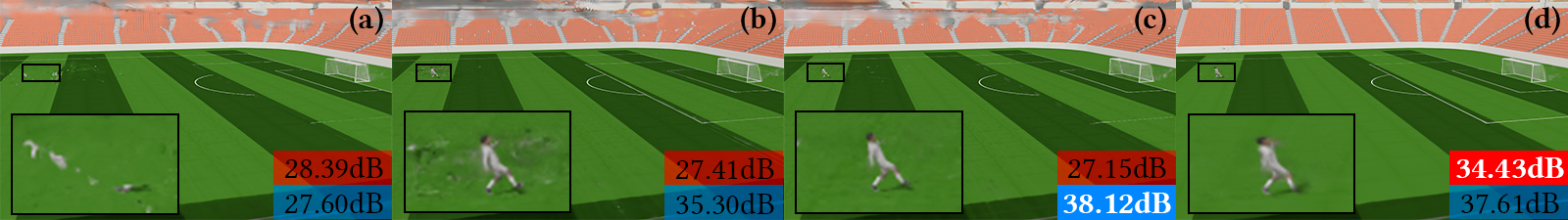}
	\caption{Illustration of \textit{focused} metrics. Each image is a prediction from the same evaluation camera pose using different model settings. The black box represents the window in which the focused metrics are computed. When using the default metrics (shown in red), only the fourth model achieves a high score, primarily due to its well-reconstructed bleachers. With the focused metrics (shown in blue), only the first model receives a low score as it fails to accurately reconstruct the dynamic content of interest.}
	\label{fig:5_4_dynmetrics}
     \Description{Horizontal grid of four images generated by four different models from the same point of view, with associated default and focused PSNR. The first image has a bad reconstruction of the bleachers and of the player, with a default PSNR equal to 28.39, and a focused PSNR of 27.60. The second image has a bad reconstruction of the bleachers, the player is well reconstructed but artifacts appear around it, with a default PSNR equal to 27.41, and a focused PSNR of 35.30. The third image has a bad reconstruction of the bleachers, but a very good reconstruction of the player, with a default PSNR equal to 27.15, and a focused PSNR of 38.12. The fourth image has a good reconstruction of the bleachers, and a slightly blurry reconstruction of the player, with a default PSNR equal to 34.43, and a focused PSNR of 37.61.}
\end{figure*}

To illustrate them, we compare \textit{default} and \textit{focused} metrics computed between one ground-truth evaluation image and novel views generated by four different versions of K-Planes, (a) to (d), where we vary the depth and width of the MLP decoder, which causes differences in prediction quality. The predicted images and associated PSNRs are depicted in Fig.~\ref{fig:5_4_dynmetrics}. The other metrics, i.e., SSIM and LPIPS, are reported in Tab.~\ref{tab:5_4_dynmetrics}. The predictions highlight the necessity for alternative evaluation methods. In (a, b, c), the bleachers are poorly reconstructed, which strongly affects all metrics, as only (d) obtains a good score. However, the player is only missing in (a), which is better reflected by the focused metrics. Also, artifacts are present around the player in (b), which underlines the need for not restricting the bounding box right around the player.

Despite these improvements, the new metrics still convey no temporal information. Furthermore, they fail if the players or the ball are not detected. For those reasons, qualitative evaluation is always preferred. With a focus on assessing the reconstruction quality of the player, we render novel viewpoints along camera paths closer to the player than the distant views used for training. In our synthetic scenes, we include additional close-up views for quantitative evaluation, which are useful when the focused metrics fail, such as in the \textit{Players} environment, in Section~\ref{wideangles}. Otherwise, one camera is excluded from the training set and used for evaluation only.

\begin{table}[!t]
\caption{Comparing default and focused metrics for the novel views shown in Fig.~\ref{fig:5_4_dynmetrics}. The default metrics are best on scenes where static elements are better reconstructed, while the dynamic-focused metrics better reflect the quality of the region of interest. Best results in bold, second-best underlined.}
\begin{tabular}{c|ccc|ccc}
    \toprule
  & \multicolumn{3}{c|}{Default}                                    & \multicolumn{3}{c}{Focused}                                  \\ 
                     & \multicolumn{1}{c}{PSNR  $\uparrow$}           & \multicolumn{1}{c}{SSIM  $\uparrow$}           & LPIPS   $\downarrow$        & \multicolumn{1}{c}{PSNR  $\uparrow$}           & \multicolumn{1}{c}{SSIM  $\uparrow$}           & LPIPS   $\downarrow$    \\ \midrule
(a)                    & \multicolumn{1}{|c}{\underline{28.39}} & \multicolumn{1}{c}{0.724} & 0.240 & \multicolumn{1}{c}{27.60} & \multicolumn{1}{c}{0.735} & 0.254 \\
(b)                    & \multicolumn{1}{|c}{27.41} & \multicolumn{1}{c}{\underline{0.745}} & \underline{0.218} & \multicolumn{1}{c}{35.30} & \multicolumn{1}{c}{0.818} & 0.079 \\
(c)                    & \multicolumn{1}{|c}{27.15} & \multicolumn{1}{c}{0.737} & 0.231 & \multicolumn{1}{c}{\textbf{38.12}} & \multicolumn{1}{c}{\underline{0.882}} & \underline{0.039} \\
(d)                   & \multicolumn{1}{|c}{\textbf{34.43}} & \multicolumn{1}{c}{\textbf{0.805}} & \textbf{0.149} & \multicolumn{1}{c}{\underline{37.61}} & \multicolumn{1}{c}{\textbf{0.924}} & \textbf{0.018} \\
    \bottomrule
\end{tabular}
\label{tab:5_4_dynmetrics}
\end{table}

\subsection{Environments}

\begin{figure}[ht!]
    \centering
    \includegraphics[width=0.8\columnwidth]{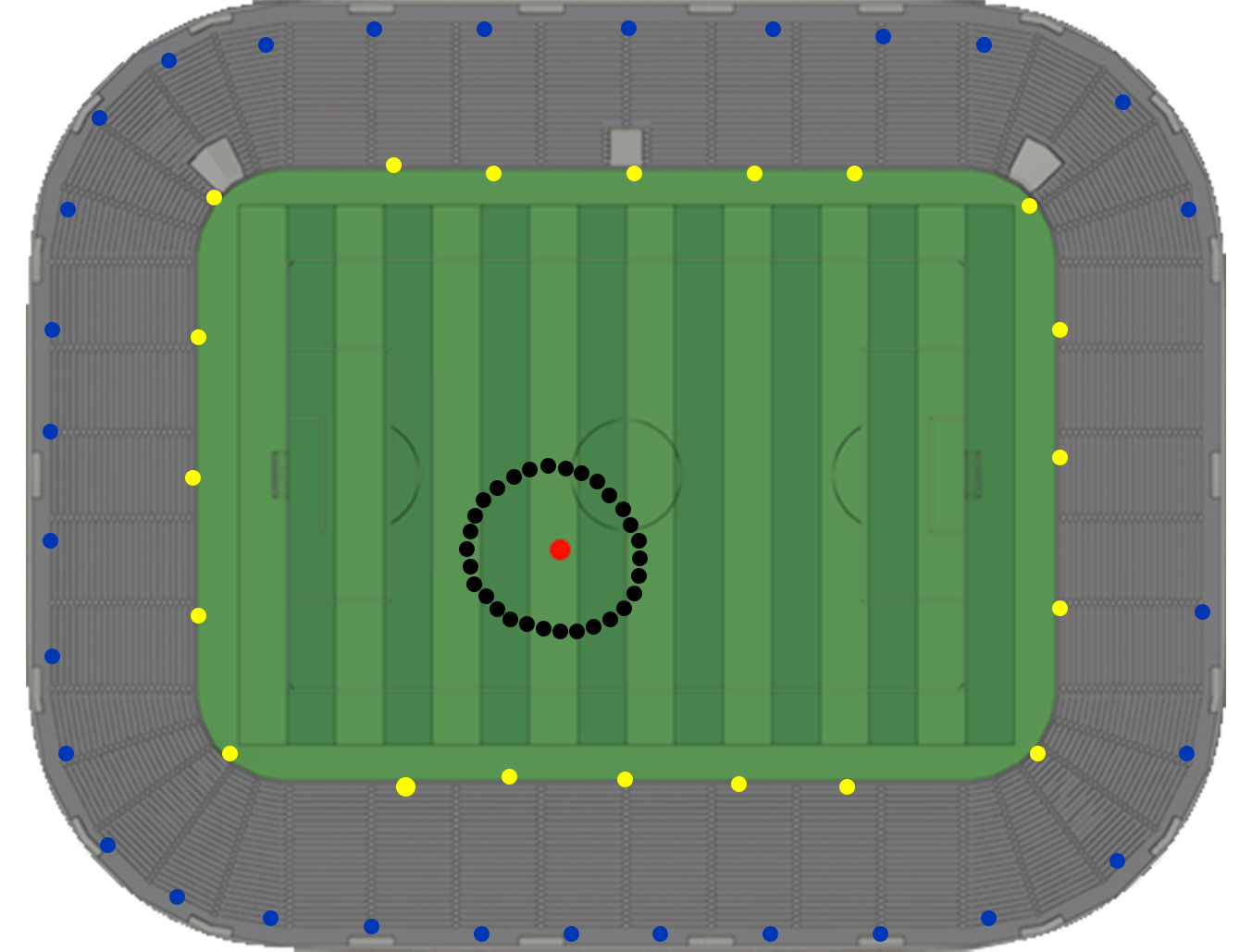}
    \caption{Illustration depicting the camera placements for each setup. The player position is highlighted in red for the \textit{Single Player} setups. The 30 close-up cameras are represented in black, the 20 broadcast-style cameras are shown in yellow, and the 30 stadium-wide angles are denoted in blue. Examples of associated training views can be observed in Fig.~\ref{fig:zoom_vs_broadcast},~\ref{fig:players_view}.}
    \label{fig:stadium_setups}
    \Description{Top-view diagram of a soccer stadium, annotated with dots that represent locations of cameras. A red dot is located near the center of the field, representing the player. It is closely surrounded by 30 black dots (first camera setup). On the border of the green field are 20 yellow dots, corresponding to the second camera setup. Finally, on the border of the bleachers, 30 blue dots correspond to the last camera setup.}
\end{figure}

\begin{figure}[ht!]
    \centering
    \includegraphics[width=\columnwidth]{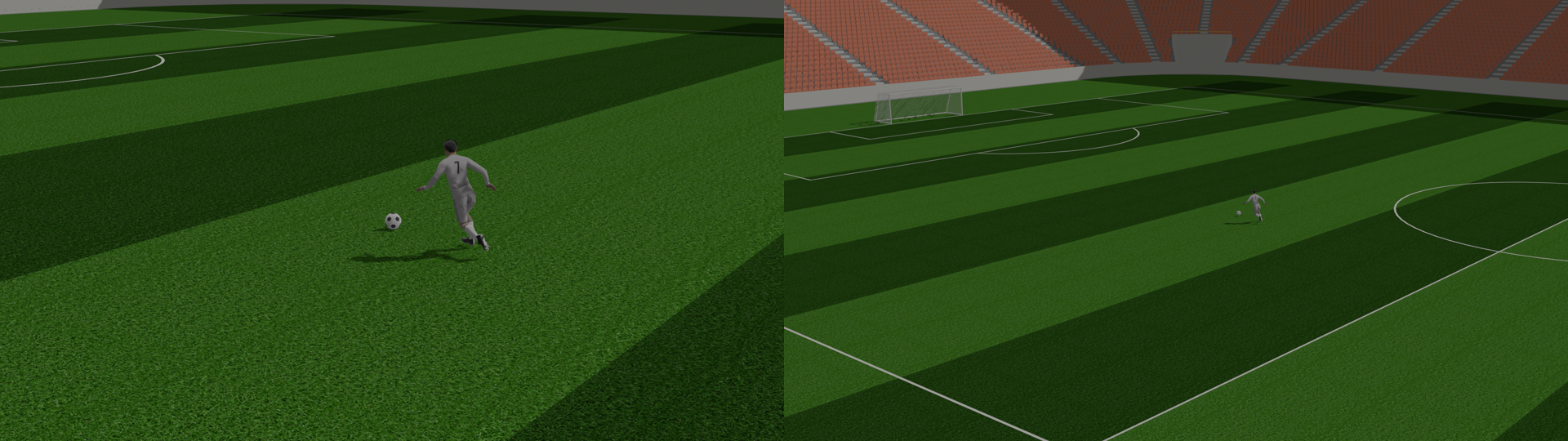}
    \caption{Camera setups for the \textit{"Single Player"} scene. Left: close-up cameras are placed around the players, similar to common datasets for novel view synthesis. Right: cameras are placed around the field and oriented toward the player, typical of broadcast conditions.}
    \label{fig:zoom_vs_broadcast}
    \Description{Two images captured by training cameras that feature a single white player on the field. On the left, a close-up view where the player is easily visible. On the right, a more distant image where the player is much smaller.}
\end{figure}

\begin{figure}[ht!]
    \centering
    \includegraphics[width=\columnwidth]{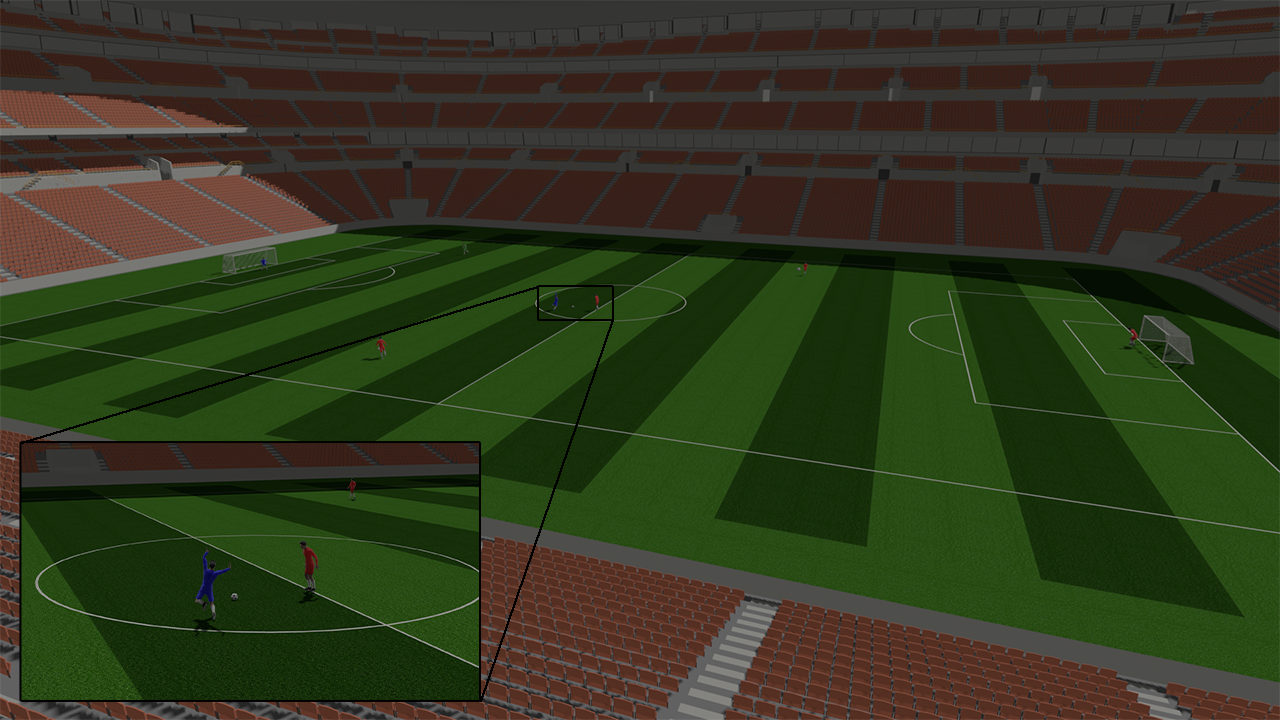}
    \caption{Example training view for the \textit{"Players"} scene, along with a close-up view dedicated for evaluation (bottom left). The players occupy a very small part of the images.}
    \label{fig:players_view}
    \Description{View of the field from a very distant camera located in the bleachers. The whole field is captured, and the players appear very small and are barely visible. A zoomed-in view of the center of the field is included.}
\end{figure}

To the best of our knowledge, most sports datasets are limited to a few synchronized cameras, and there are no public datasets that include dozens of synchronized and calibrated views. For example, the KTH Multiview Football Dataset II \cite{MultiviewBodyPart_kazemi2013} only contains three synchronized cameras, that often focus on a single player.

In this exploratory work, we build synthetic environments to circumvent the lack of real data. 
The scenes are of increasing complexity, starting from relatively close-up views, commonly used with NeRFs, then using more distant cameras giving a field of view similar to what is used in broadcast coverage, and finally considering even more distant cameras placed in the bleachers covering the whole field with more players. The different setups are illustrated in Fig.~\ref{fig:stadium_setups}. This allows us to progressively tackle the challenges that occur with soccer environments, mainly the reconstruction of small dynamic content. All cameras are static and the environments are built using Blender \cite{Blender3DModelling_community2018} with player models from Adobe Mixamo \cite{Mixamo}, and a stadium model available under a free CC0 license \cite{Stadium}.

\paragraph{Single Player: Close-up Views} 
This synthetic environment features a single player placed at the center of the field, shooting a ball. This first camera setup is composed of 30 close-up views around the player and resembles typical conditions of benchmarks like DyNeRF \cite{Neural3DVideo_li2022}. An example training image is depicted in Fig.~\ref{fig:zoom_vs_broadcast} (left).

\paragraph{Single Player: Broadcast-style Views}
Within the same environment, we consider a second camera configuration that features 20 views placed around the field, whose field of view is close to broadcast conditions. The player represents only a tiny portion of the images. An example training image is depicted in Fig.~\ref{fig:zoom_vs_broadcast} (right).

\paragraph{Players: Stadium-wide Views}
This more complex environment features several players and balls interacting all over the field, captured by 30 wide-angle cameras placed high up in the bleachers and are thus much more distant from the field. Six additional cameras, used exclusively for evaluation, are placed near the players for more meaningful results. In this setup, training views cover the whole field at all times but cover very few details about the players and balls due to their large distance. An example training view is depicted in Fig.~\ref{fig:players_view}.

\section{Experiments}\label{experiments}

In this section, we assess the performance of K-Planes \cite{KPlanesExplicitRadiance_fridovich-keil2023} and NeRFPlayer \cite{NeRFPlayerStreamableDynamic_song2023} in increasingly complex environments, each described in Section~\ref{implem}. 

\subsection{Single Player: Close-up Views}

As a first attempt, we run the original models from the initial papers in similar conditions to traditional datasets \cite{Neural3DVideo_li2022, broxton2020immersive}. The player occupies a large region of the training images, is captured by 30 cameras, and performs smooth motion. In these settings, the models are able to reconstruct the stadium flawlessly. The player's motion is reconstructed, but the texture is blurry, even when using larger models. The ball is not reconstructed when moving fast in the air and disappears.

\begin{figure*}[ht!]
    \centering
    \includegraphics[width=\textwidth]{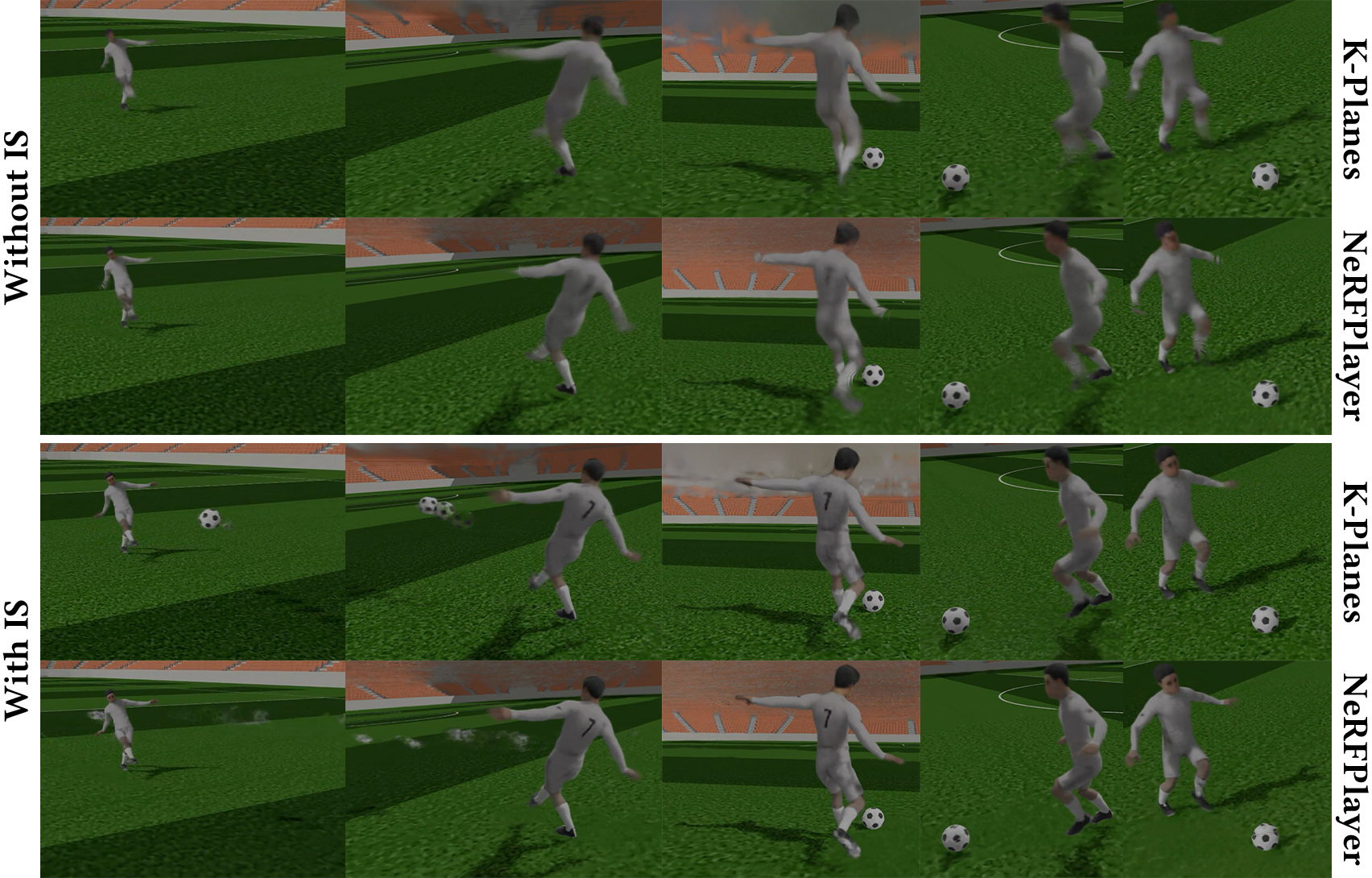}
    \caption{Comparative results between K-Planes and NeRFPlayer on the \textit{Single Player} environment with \textit{close-up cameras}, with and without ray importance sampling (\textit{IS}). Overall, both models obtain similar results. NeRFPlayer tends to recover slightly more details on the player, possibly due to the factorization of K-Planes, but induces more artifacts and does not manage to reconstruct the ball when in the air. Importance sampling drastically improves results for both models and allows K-Planes to recover the ball when in the air.}
    \Description{Grid of image results generated from novel positions. The first two rows are without ray importance sampling, the last two rows are with it. The first and third rows use the K-Planes model, and the second and fourth rows use NeRFPlayer. Each image depicts a player, which with better or worse quality depending on the settings. In the first two rows, the texture of the player is blurry but the environment is well reconstructed. With importance sampling (last two rows), the results are much more clear and well-defined. There is very few differences between the two models (between row 1 and 2, and row 3 and 4).}
    \label{fig:kp_vs_np}
\end{figure*}

We can circumvent these issues by employing an improved pixel sampling strategy. Traditionally, training rays are traced by uniformly sampling pixels although dynamic content, especially if small, should be sampled more often. In \cite{Neural3DVideo_li2022}, several improved strategies are described, known as \textit{Ray Importance Sampling} (IS). This new sampling strategy, which prioritizes sampling dynamic content pixels, is particularly necessary for setups like ours, considering the scale of dynamic objects, even in this first more ideal environment. This general modification, which can be applied to both models, yields substantial improvements in quality and training time. Renderings are performed around the player with both models, with and without ray importance sampling, and are depicted in Fig.~\ref{fig:kp_vs_np}. Associated metrics, computed using a dedicated evaluation camera, are reported in Tab.~\ref{tab:kp_vsnp}. Visual details are recovered much quicker, and final results are drastically more detailed when using importance sampling. Overall, results are similar between NeRFPlayer and K-Planes when using similar model sizes, as depicted in Fig.~\ref{fig:kp_vs_np}. NeRFPlayer tends to recover slightly more details on the player but produces more artifacts around it. While it is not able to reconstruct the ball when it is in the air, K-Planes manages to reconstruct it, although ghosting effects appear. When not using importance sampling, the ball is never reconstructed. Here, the use of focused metrics barely affects our interpretation of the results, due to the player's scale in the images, which causes the bounding boxes to cover a large part of the view. While the focused PSNR improves when using importance sampling, the other metrics sometimes degrade, which does not support qualitative results from Fig.~\ref{fig:kp_vs_np}. This may be explained by the fact that IS helps to partially reconstruct the ball, which introduces artifacts.

\subsection{Single Player: Broadcast-style Views}

While the models perform well with close-up cameras, such views are usually not available in practical applications. Here, we experiment with the same scene but observed by more distant training views, which are positioned like broadcast cameras, all around the field.

\begin{figure*}[ht!]
    \centering
    \includegraphics[width=\textwidth]{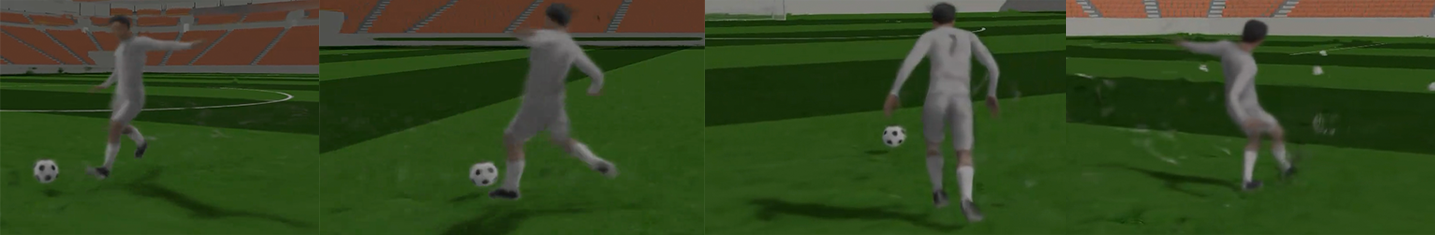}
    \caption{Novel views synthesis in the \textit{Single Player} environment using a camera setup similar to broadcast conditions. Results are obtained from K-Planes with Ray Importance Sampling. The player is well reconstructed, but its texture is blurrier when compared to using closer cameras. Despite the utilization of Importance Sampling, the model fails to accurately reconstruct the ball in motion (on the right).}
    \label{fig:broadcast_views}
    \Description{Grid of four images generated from novel positions of a synthetic soccer player. Although slightly blurry, the player is well reconstructed, its texture is quite detailed and the environment is also well visible. The fourth image depicts artifacts as the ball is not reconstructed.}
\end{figure*}

Example renderings, using ray importance sampling, are depicted in Fig.~\ref{fig:broadcast_views}. The player is still reconstructed accurately, although less detailed, compared to the closer camera setup. In these new conditions, importance sampling is even more necessary, as the player is barely reconstructed without it. However, even with IS, the ball is not reconstructed when in motion. Instead, artifacts appear everywhere in the direction of cameras.

\subsection{Players: Stadium-wide Views}\label{wideangles}

This final synthetic environment moves further away from the center of the scene and features 30 wide-angle cameras, located high in the bleachers, that cover the whole field. Many players are present on the soccer pitch, interacting with each other and with balls. This setting is particularly challenging, due to the very small visibility of players in the training images.

\begin{figure*}[ht!]
    \centering
    \includegraphics[width=\textwidth]{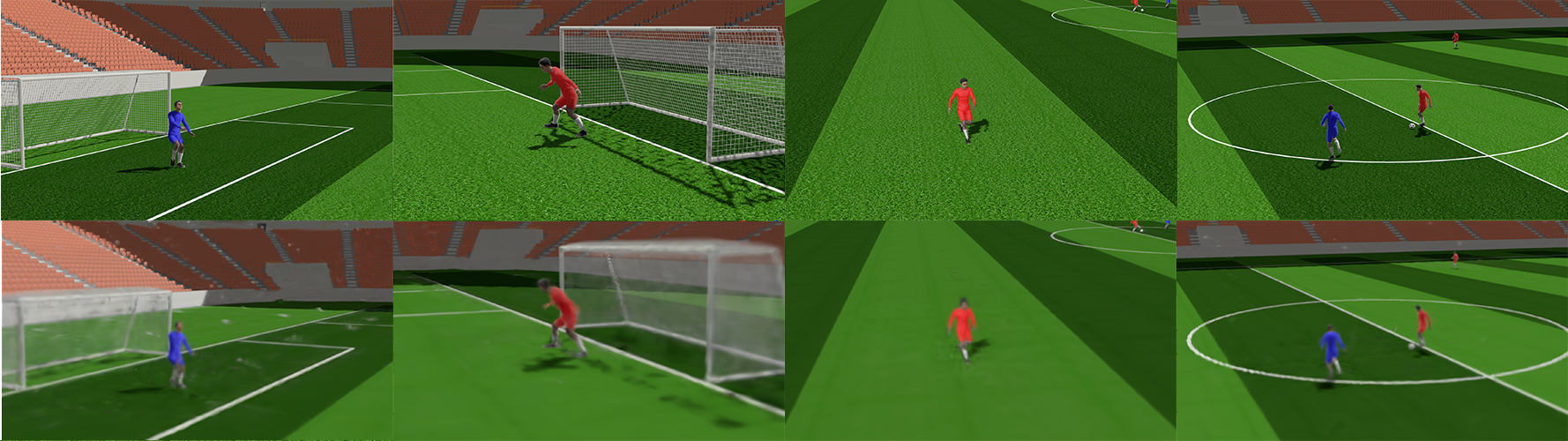}
    \caption{Ground truths (top) and predictions (bottom) for the \textit{Players} environment from close-up views dedicated to evaluation. In these difficult conditions, the players are still reconstructed, although quite blurry.}
    \label{fig:wideangle_1}
    \Description{Grid of ground truths and novel views in a synthetic environment with several players. Each row contains four images. Each image corresponds to a different location and action on the field: (1,2) Both goal keepers, (3) one player walking, and (4) two players interacting with the ball at the center of the field.The first row contains true views of the synthetic scene. The second row contains associated generated views, depicting blurry players which are nonetheless discernable.}
\end{figure*}

\begin{figure*}[ht!]
    \centering
    \includegraphics[width=\textwidth]{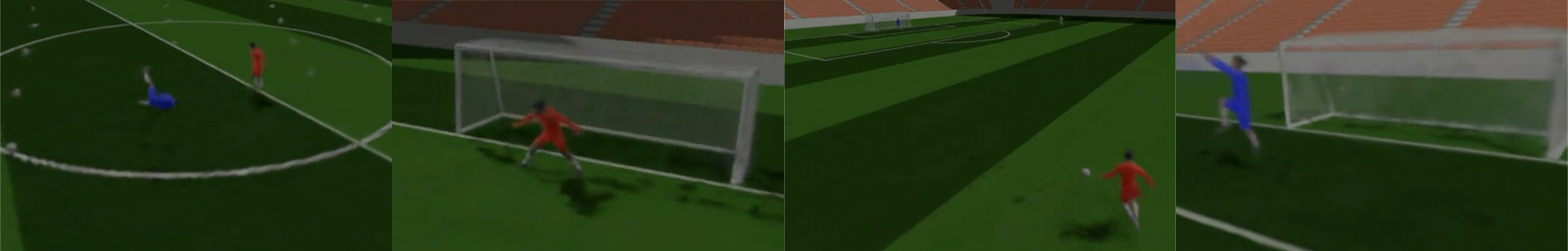}
    \caption{Additional novel views synthesis results in the \textit{Players} environment. In these challenging conditions, the motion and position of the players are correctly reconstructed, albeit significantly blurry. The ball is not reconstructed and causes artifacts visible on the whole field (leftmost image).}
    \label{fig:wideangle_2}
    \Description{Grid of four novel views in a single row in a synthetic soccer environment. First image: two players interacting at the center of the field, they are reconstructed by the model, but quite blurry, and the ball is missing. Second image: a red goal keeper is quite well reconstructed although still blurry. Third image: a red soccer player about to shoot, the static ball is reconstructed. Fourth image: a blue goal keeper jumps to catch a ball and is - again - reconstructed but blurry.}
\end{figure*}

Results are depicted in Fig.~\ref{fig:wideangle_1} and \ref{fig:wideangle_2}. Even in this very challenging configuration, the players are reconstructed and we can distinguish their motion. However, even with larger models, 1080p training images, and ray importance sampling, the results remain blurry. The ball is barely reconstructed when moving slowly, and not at all when moving fast (e.g., when being shot). Such camera setups, therefore, seem to be limited for detailed results, at least when using no domain knowledge.

\begin{table*}[ht!]
\caption{Quantitative results for both models with and without ray importance sampling (see Fig.~\ref{fig:kp_vs_np}). Due to using closer cameras, the focused metrics have a limited impact on the results.}
\centering
\begin{tabular}{cl|ccc|ccc}
\toprule
 &  & \multicolumn{3}{c|}{Default} & \multicolumn{3}{c}{Focused} \\ \cline{3-8} 
 &  & PSNR  $\uparrow$ & SSIM  $\uparrow$ & LPIPS $\downarrow$ & PSNR  $\uparrow$ & SSIM  $\uparrow$ & LPIPS   $\downarrow$ \\ \midrule
\multicolumn{1}{c|}{\multirow{2}{*}{\begin{tabular}[c]{@{}c@{}}Without\\ Importance Sampling\end{tabular}}} & K-Planes & \textbf{32.84} & \textbf{0.786} & \textbf{0.167} & 34.41 & \textbf{0.816} & \textbf{0.126} \\
\multicolumn{1}{c|}{} & NeRFPlayer & 31.54 & 0.754 & 0.211 & 34.55 & 0.807 & 0.181 \\ \hline
\multicolumn{1}{c|}{\multirow{2}{*}{\begin{tabular}[c]{@{}c@{}}With\\ Importance Sampling\end{tabular}}} & K-Planes & 32.53 & 0.751 & 0.199 & \textbf{35.06} & 0.788 & 0.156 \\
\multicolumn{1}{c|}{} & NeRFPlayer & 31.26 & 0.721 & 0.225 & 34.78 & 0.781 & 0.191 \\ \bottomrule
\end{tabular}
\label{tab:kp_vsnp}
\end{table*}

\section{Discussion}\label{discussion}

In this exploratory work, we compared recent state-of-the-art dynamic NeRF models, i.e., K-Planes \cite{KPlanesExplicitRadiance_fridovich-keil2023} and NeRFPlayer \cite{NeRFPlayerStreamableDynamic_song2023}, in increasingly complex soccer environments, to assess their readiness for broadcast-quality novel-view video synthesis of soccer replays. In the ideal NeRF setup, where close-up cameras capture detailed views of the target moving objects, the models reached great reconstruction quality. However, when using distant views in a camera setup similar to the best-result proprietary system \cite{IntelTrueView}, the results offered by general dynamic NeRFs drastically degrade. In such distant camera setups, we showed that incorporating additional components to the original models, like ray importance sampling \cite{Neural3DVideo_li2022}, becomes an absolute necessity. 

We tried to avoid working with very high-resolution images, as opposed to \cite{IntelTrueView}, limiting our input image data to 1080p. Indeed, even though increasing the image resolution is an obvious path of improvement toward capturing fine details, the massive amounts of data thus generated are close to being prohibitive computationally, and we wanted to explore more economical solutions. For similar reasons, we avoided resorting to domain-specific priors in this study, as such priors can be arduous and costly to produce, e.g., an accurate skeletal reconstruction of the players. Assuming such restrictions, and despite our improving components, we must conclude that general dynamic NeRF models may fall short of meeting the high-quality requirements of the broadcast industry for novel view synthesis of soccer replays.

Although it was not the focus of our work, another inconvenience of using dynamic NeRFs in broadcast applications might be their time performance. Indeed, the models we selected require one hour to train on thirty 4-second clips and 5 minutes to render a 10-second video (about 1FPS for 1080p rendering). However, we believe that training times could very certainly be lowered, notably by pre-training a model for the empty stadium. Nonetheless, even the most recent models \cite{MixedNeuralVoxels_wang2022} require more than 15 minutes of training, which while being unsuitable for live replay, might fit post-match applications.

Still, we believe that dynamic NeRFs could play an important role as the core part of a fully satisfying solution. Following the same line of work as what was done in our study, a first path of improvement would be to try incorporating other general components into dynamic NeRFs. The visibility loss from Nerfbusters \cite{NerfbustersRemovingGhostly_warburg2023}, the improved proposal sampling from Zip-NeRF \cite{ZipNeRFAntialiasedGridbased_barron2023}, and the restorer from NeRFLiX \cite{NeRFLiXHighqualityNeural_zhou2023} are promising components that would certainly be beneficial to a detailed reconstruction of soccer scenes in distant camera setups. Nevertheless, using such general improving components may still not be enough for the task.

Although using absolutely no domain knowledge is appealing, it may be necessary to use some domain-specific components to reach broadcast-quality results as well as a better time performance during training, more in line with broadcast time constraints. Yet, one should be cautious of the complexity and costs associated with bringing specific models within a solution. For instance, while showing impressive results in controlled working conditions, NeRFs that focus on human reconstruction \cite{HumanNeRFEfficientlyGenerated_zhao2022, HumanNeRFFreeviewpointRendering_weng2022, HOSNeRFDynamicHumanobjectscene_liu2023} are not directly usable with a distant camera setup such as ours, and would require considerable adaptation to reconstruct humans in more diverse, less-constrained configurations, such as multiple humans at arbitrary positions.

As manifest as increasing the input image resolution, another path of improvement is to obtain and leverage more zoomed-in input views, together with the distant views given by our chosen camera setup inspired by \cite{IntelTrueView}. Our study indeed showed that broadcast-style views may capture enough details to render novel views with near-acceptable quality for the target application. Such cameras could not be static, though, and it is unrealistic to suggest manning dozens of additional broadcast-style cameras with operators tasked to follow the action. This naturally leads to consider using the image data coming from the actual broadcast cameras, which are used to cover the soccer event. Including broadcast moving cameras within the reconstruction task would introduce new difficulties, such as motion blur, less accurate camera calibration, view sparsity for the zoomed-in region of interest, and the inadequacy of importance sampling as it relies on static cameras. The benefits could however outweigh the difficulties. First, robust SfM tools could still be used with satisfaction in a mixed setup of distant static cameras and broadcast moving cameras, to retrieve the calibration of the moving cameras at all times. Second, using such a mixed setup could allow using less static cameras than the dense 20-30 camera array considered in this study. A case could even be made that broadcast cameras become the main source of information in an economical solution, using all available NeRF extensions that deal with sparse camera setups, such as depth supervision \cite{DepthSupervisedNeRFFewer_deng2022} based on what SfM tools output for the scene structure, along with the camera calibrations.

An indirect, but very important path of improvement is the design of better evaluation metrics for dynamic NeRFs. Evaluating these models in less-frequently considered dynamic environments, such as soccer, poses significant challenges. In our study, we proposed a simple yet better method for computing evaluation metrics. However, much more could be made, particularly in detecting general moving content, incorporating temporal information, and finding ways to accurately reflect the challenging reconstruction quality of the ball. Proper evaluation of these models is \textit{crucial} because, without accurate assessment, it is difficult to determine the readiness of a method for real-world applications.

Finally, also an indirect path of improvement: acknowledging and remedying the lack of public multi-view soccer datasets. Even a single image dataset of a dozen synchronized cameras capturing a few soccer actions would be of tremendous interest to the community. The synthetic environments we built are a modest proxy of such a dataset, that we publicly release along with all the code used for the experiments, both ready to use in Nerfstudio, an open-source framework for NeRF research. We strongly encourage building richer datasets, both by extending our scenes and by recording real data using enough synchronized and calibrated cameras.

\begin{acks}
We sincerely thank EVS Broadcast Equipment for providing the necessary compute for the various conducted experiments. We also thank the Nerfstudio community for helpful insights. 
\end{acks}

\bibliographystyle{ACM-Reference-Format}
\balance
\bibliography{paper}

\end{document}